\documentclass[11pt]{article}
\usepackage{geometry}
\usepackage{coling2020}
\usepackage{times}
\usepackage{makecell} 
\usepackage[detect-all]{siunitx}

\usepackage{url}
\usepackage{latexsym}
\usepackage{microtype}
\usepackage{csquotes}
\usepackage{tabularx,xcolor}
\usepackage{amssymb}
\usepackage{amsmath}
\usepackage{latexsym}
\DeclareMathOperator*{\argmin}{arg\,min}

\colingfinalcopy 

\title{BUT-FIT at SemEval-2020 Task 4: Multilingual commonsense}

\author{Josef Jon, Martin Fajcik, Martin Docekal, Pavel Smrz\\
  Brno University of Technology, Faculty of Information Technology\\
  612\,66 Brno, Czech Republic \\
  {\tt \{ifajcik,ijon,idocekal,smrz\}@fit.vutbr.cz} }

\date{}

\begin{document}
\maketitle
\begin{abstract} 
This paper describes work of the BUT-FIT's team at SemEval 2020 Task 4 - Commonsense Validation and Explanation. We participated in all three subtasks. In subtasks A and B, our submissions are based on pretrained language representation models (namely ALBERT) and data augmentation. We experimented with solving the task for another language, Czech, by means of multilingual models and machine translated dataset, or translated model inputs. We show that with a strong machine translation system, our system can be used in another language with a small accuracy loss. In subtask C, our submission, which is based on pretrained sequence-to-sequence model (BART), ranked 1st in BLEU score ranking, however, we show that the correlation between BLEU and human evaluation, in which our  submission ended up 4th, is low. We analyse the metrics used in the evaluation and we propose an additional score based on model from subtask B, which correlates well with our manual ranking, as well as reranking method based on the same principle. We performed an error and dataset analysis for all subtasks and we present our findings.

\end{abstract}

\blfootnote{
    %
    %
    %
    %
    %
    %
     \hspace{-0.65cm}  
    This work is licensed under a Creative Commons 
    Attribution 4.0 International License.
    License details:
    \url{http://creativecommons.org/licenses/by/4.0/}.
}

\section{Introduction}
\label{intro}

Commonsense knowledge is collection of facts, that an average human being is expected to know and be able to reason with these facts. It includes the ability to asses physical qualities, behavior and purpose of inanimate objects, animals and people. In other words, it is an implicit knowledge, something we feel we do not have to explain.  Since this topic is among the most important challenges in artificial intelligence, there is a large number of publications regarding commonsense -- see for example books by \newcite{mueller2014commonsense} or \newcite{davis2014representations}.
 
The goal of SemEval 2020 Task 4 - Commonsense Validation and Explanation \cite{wang-etal-2020-semeval} was to asses how current NLP approaches compare with humans in terms of such knowledge. Three subtasks were devised to determine the ability of the participating systems to reason within commonsense knowledge in English. 

In the first subtask, the goal was to differentiate between statements that make sense and those that do not. In the second and third subtask, only nonsensical statements were considered, and the assignment was to explain why are the statements against common sense. In subtask number two, the systems were presented with three possible choices for the explanation, while in the third subtask, the systems were expected to generate the explanation from scratch. A more detailed description of the tasks is provided in the task paper.

Our submissions are based on recent pretrained models, namely ALBERT \cite{lan2019}  and BART \cite{lewis2019bart}. We experimented with data augmentation using round trip translations of the datasets and also with machine translation of either the queries, or the whole dataset into another language (Czech). Our code is available in a github repository at \url{https://github.com/cepin19/semeval2020_task4} .
\section{System overview}
\subsection{Language Representation Models}
Recently, pretrained language representation models (LRMs) became state-of-the art in many natural language processing tasks. We have experimented with following LRMs:

\textbf{BERT} \cite{devlin2019} or Bidirectional Encoder Representations from Transformers is a method of pretraining language representations on large amounts of unlabeled text and a class of models trained by this method. BERT is a Transformer \cite{vaswani2017attention} model trained with two objectives -- Masked Language Modelling (MLM) and Next sentence prediction (NSP). For the MLM objective, 15\% of the input tokens are replaced by \texttt{[MASK]} token, creating a noised version of the input. The model denoises the input by predicting the original tokens in the masked positions. In NSP training, the model predicts whether two sentences from the training data follow each other.

\textbf{RoBERTa} \cite{liu2019} is based on an analysis of design choices in BERT. It changes the tokenization method from WordPiece to byte-pair encoding \cite{sennrich2016}, masks the input sequences dynamically during the training, removes the next sentence prediction objective, trains on larger batches formed by longer sequences, on more data and for a longer time.

\textbf{ALBERT} by \newcite{lan2019} introduces two modifications to decrease the number of parameters, while not negatively affecting the accuracy. Firstly, the embedding matrix is factorized into two matrices, one for context-independent vocabulary word embedding, and one for context-dependent representation, which is processed by the hidden layers. This allows for increase of the hidden layer size without increasing the vocabulary embedding size, which is beneficial since a larger hidden layer size has greater positive impact on model's accuracy than a larger embedding size. Secondly, some layers of the model share parameters.



Both of these adjustments slightly harm the model accuracy. However, due to them, number of parameters is greatly decreased. This allows creation of much larger models within the same memory restrictions, which leads to accuracy increase, higher than the decrease caused by the adjustments. 

\textbf{GPT-2} \cite{radford2019language} is a Transformer language model, with up to 1.5B parameters in its largest variant, trained on 40GB of text with the next word prediction objective.

\subsection{Sequence-to-sequence models}
The last subtask can be framed as a sequence-to-sequence problem. We employed Transformer-based encoder-decoder models to generate the target sentence conditioned on the source sentence.

\textbf{Transformer} \cite{vaswani2017attention} model is the foundation of all the models presented in this section. The most important feature of the model is the attention mechanism, which is applied to all input symbols repeatedly. In each step, all symbol representations are weighted to create a new representation for the given symbol. Since all the source symbols are processed at once, the model is more parallelizable than RNNs, which processes the input sequentially. The non-sequential nature of the model also helps with modelling long-range dependencies.  

\textbf{BART} \cite{lewis2019bart} is a Transformer model pretrained with denoising autoencoding objective. The training data are corrupted by a noising function, which can in theory be arbitrary, and the model learns to reconstruct the original text. The published pretrained models are trained with a text filling objective, where spans of tokens in the source are replaced by a single \texttt{[MASK]} token. 30\% of the input tokens are replaced and span lengths are sampled for in Poisson distribution with $\lambda=3$. This objective is harder than replacing single tokens with \texttt{[MASK]}, since the model does not know the length of the replaced span. On top of this type of corruption, all sentences in training documents are randomly permuted and the model needs to learn to reorder them correctly.

\subsection{Multilingual systems}
Aside from attempting to develop a system for English, we focused on ways to solve the task also for the Czech language, using machine translation. We consider two approaches. First, translating the training data and training a model directly for Czech language. The second approach is to use a model trained on English data, and only translate the inputs, given in another language, into English. Both of these approaches have drawbacks. Using the first method, two problems arise -- pretrained LRMs for other languages are not on par with the English models, and the machine translation of the dataset contains errors. We mitigate the second problem by choosing a high-resource language pair and strong neural machine translation (NMT) models.

The second approach suffers from  similar translation quality problem, which is further aggravated by the fact that the data are translated twice, which can lead to error propagation -- first, the validation set is translated into a Czech and then back into English, to simulate Czech input. On the other hand, the sentences translated into Czech are in fact English-Czech translationese, meaning they still retain some characteristics of English, and it might be easier for the NMT system to translate them back into English, compared with real Czech sentences (for more details on translationese in NMT see for example \newcite{toral2018attaining}).


\subsection{Subtask A}
In subtask A, the goal is to determine which of two statements is more against common sense. We tuned the pretrained LRMs described above, namely BERT-base, roBERTa-base, roBERTa-large, ALBERT-xxlarge and multilingual BERT on the training data using cross-entropy minimization objective. The two statements are delimited by \texttt{[SEP]} token or the equivalent depending on the model, e.g. for BERT the input is in form \texttt{[CLS]statement1[SEP]statement2[SEP]}. The statements are classified based on a linear transformation of the \texttt{[CLS]}-level output, after applying dropout. 
\subsection{Subtask B}
In this task, the system is given a statement that is against common sense, and three explanations why, as an input. The system selects one of the three explanations. Each of the options is encoded separately in similar fashion as in subtask A, i.e. for BERT \texttt{[CLS]statement[SEP]reason\_i[SEP]}, where \texttt{statement} is the nonsensical statement, $i=\{1,2,3\}$ and \texttt{reason\_i} is one possible explanation. The model is run for each of these three inputs, and the outputs of the last hidden layer are pooled together. A linear transformation and softmax are applied to the pooled outputs to obtain class probabilities.

Another approach we have experimented with is to compute perplexity of each option with respect to the input using a generative model from subtask C. We discuss this approach more in depth in Section \ref{sec:results}.

\subsection{Subtask C}
We approached the task with two slightly different ways. First, as a sequence-to-sequence task, where the nonsensical statement is source sequence and the explanation is target sequence. We experimented with the vanilla Transformer model and BART \cite{lewis2019bart}. The explanation is generated by searching for most probable symbol sequence using beam search during inference.

And second, as a language generation task, where the nonsensical statement is used as a prompt for a generative language model. For this approach, the statement and the explanation are concatenated together, and a language model (GPT-2 in our case) is trained on these  sequences with a next token prediction objective. At the test time, only the statement is known, and based on the statement, the model recursively generates tokens from the explanation until end-of-sentence token is generated. We used greedy decoding for this approach, choosing only the most probable token at each step.

As a baseline, we computed the BLEU score of the input against the reference. Compared to other approaches, the resulting score is the second best. This is caused by the fact that the input sentence and the correct explanation have a large token overlap. The metric is further discussed in the analysis.

We also used the model from the previous subtask to rerank explanations generated by the different models we have experimented with. This approach is described more in depth in Section \ref{sec:results}.
\newpage
\section{Experimental setup}
\subsection{Data}
The organizers provide data split into training/validation/test sets, comprising of 10000/997/1000 examples respectively. The label distribution for subtasks A and B is well balanced -- 49.8\% to 50.2\% in subtask A, and 32\%, 33.6\% and 34.4\% in subtask B. 

\subsection{Tools}
Our experiments with \texttt{BERT}, \texttt{roBERTa}, \texttt{ALBERT} and \texttt{GPT-2} models were based on \texttt{transformers}\footnote{https://huggingface.co/transformers/} \cite{HuggingFacesTS} library written in PyTorch \cite{NEURIPS2019_9015}. We used \texttt{Marian} \cite{mariannmt} with \texttt{subword-nmt} BPE \cite{sennrich2016} for  experiments with the vanilla Transformer. Finally, for \texttt{BART}, we utilized the original implementation in \texttt{fairseq} \cite{ott2019fairseq}.

\subsection{Translation}
English to Czech translation was performed by a strong Transformer model (29.5 BLEU on WMT19 \cite{ws-2019-machine-translation} test set), trained on CzEng 2.0 corpus\footnote{\url{http://ufal.mff.cuni.cz/czeng}}, consisting of 61M lines of parallel data and 51M lines of synthetic backtranslated data. Most of the training sentences were translated correctly. For instance, out of the first 100 examples from subtask A (200 sentences), only 2 sentences were translated incorrectly and 4 translations had minor fluency issues. Examples of translations are shown in Supplemental material, section \ref{translation_examples}. For some of the experiments, Czech sentences were translated back into English, using a smaller, also Transformer-based, Czech to English NMT model.

Some of the sentences in the data are completely in uppercase. Such sentences were lowercased before the translation. Missing punctuation was added at sentence endings. Our round trip translated data had BLEU score of 55.85 when compared with the original data. 

To asses the influence of translation quality, we also translated the data by a popular online tool\footnote{\url{https://translate.google.com}}, which in our experience, in most cases, provides lower quality translations than the model described above (23.9 BLEU on WMT19 test set). The round trip translations created by this tool had BLEU of 30.65 when compared with the original data, showing that these translations are less similar to the original sentences. 

\subsection{Data augmentation}
We experimented with augmenting the training data by paraphrasing the statements. As a simple approach to paraphrasing, we used round trip translations described above, originally created for the multilingual experiments.  We concatenated these modified examples with the original training data, deduplicating the examples which were identical after the round trip translation. On the one hand, we noticed that there are very rarely any mistakes in the translations. On the other hand, the translations are very similar to the original training data and more diverse paraphrases, e.g. ones created by LRMs trained on paraphrasing datasets, could have better effect as an augmentation method.

\section{Results and analysis}
\label{sec:results}
In this section, the results of models described earlier are presented, along with  data and error analysis.
\subsection{Subtask A}
We experimented with various LRMs in this task, both in English and in Czech. We used round-trip translated data as an augmentation technique. Accuracy and F1 scores of the systems are presented in Table \ref{tab:subtaskA_results}. 

The results follow a general pattern in recent LRMs' performance -- in a large portion of tasks, roBERTa provides better results than BERT, and ALBERT-xxlarge outperforms roBERTa, even with smaller number of parameters.

Our final submission is an ensemble of ALBERT-xxlarge models. It  ranked 7th in the final ranking, 1.2\% accuracy behind the winning submission. To choose the models to use in the ensemble, we saved output probabilities of 12 checkpoints that had the highest F1 scores on the validation data of all runs. All of the possible combinations of models were generated, and the output probabilities were averaged over all the models in the given combination. The final ensemble is formed by 5 models, 3 of which were trained on the original data,  one both on original and round trip translated data, and one solely on the round trip translated data. This suggests that even though training on augmented data did not help for a single model, it increases the diversity of the predictions and thus helps in ensembled performance.

Our best ensemble made 28 errors on the validation set. We present breakdown of these errors and comments in Supplemental material, section \ref{subtaskA_examples}.

\begin{table}[t!]
\begin{center}
\begin{tabular}{|l|cccc|}
\hline \bf Model &\bf dev accuracy  & \bf dev F1 & \bf test accuracy   & $\boldsymbol{\#\theta}$ \ \\ \hline
Majority & 50.21 &- & - &  - \\ \hline
BERT-base &  83.55 & 83.13 & 81.2 & 110M \\ \hline
mBERT cs &  69.31	& 71.08 & 70.3  & 110M \\  \hline
mBERT en  &77.93&	78.05 &  77.5 &  110M\\  \hline
mBERT RTT dev  &  78.34 & 77.55 & 75.6 &  110M\\  \hline
mBERT RTT dev+train &  75.53 & 75.2  & 75.9 &110M \\  \hline

roBERTa-base &   86.86 & 86.37  & 85.9& 125M\\  \hline
roBERTa-large & 93.58& 93.39 & 91.7 &  355M \\ \hline

ALBERT-xxlarge   & 96.39 & 96.23 & 95.4 &   223M\\ \hline
ALBERT-xxlarge + synth  & 96.29 &  96.11  & 95.4  & 223M  \\ \hline
ALBERT-xxlarge RTT dev worse &  91.68 &  91.14 & 91.6  & 223 \\ \hline

ALBERT-xxlarge RTT dev &  93.78 & 93.45 & 92.1 & 223M \\ \hline
ALBERT-xxlarge-ens* & \textbf{ 97.29}  & \textbf{97.13} & \textbf{95.8}  & 5*223M\\ \hline

\hline
\end{tabular}
\label{tab:subtaskA_results}
\caption{Results of subtask A, accuracy and F1 scores on validation and test sets. Column $\#\theta$ shows the number of parameter of each model.  Cs after the model name means both the training and dev data were translated automatically to Czech and the model was trained on Czech data. RTT dev means the model was trained on the original data, but the dev and test sets were translated by our NMT models into Czech and back into English to simulate a Czech input. For RTT dev worse, the translations were carried out by a popular online translation tool. RTT dev+train denotes a model that was trained on data translated to Czech and back into English, and evaluated using equally preprocessed dev set. Synth data mean round trip translated train sets, which were concatenated with the original train sets. Our final submission is marked with an asterisk. }
\end{center}
\end{table}

\paragraph{Multilingual setting} To experiment with the multilingual setting, we used a multilingual BERT model. As a baseline, the pretrained model model was fine-tuned on the original data. Subsequently, the training data were translated into Czech by our NMT system and we fine-tuned the pretrained model on the translated dataset. Czech is among the languages used during the pretraining. The Czech version lags behind the English one by 7 F1 points. This is a large difference, considering that the quality of the translations seems to be very good. 

We ran another experiment, using a model trained on the original English data, but translating the validation set to Czech and back to English, to simulate a use case where the model is used on Czech sentences, which are then machine translated into English. In this setting, the validation set F1 score is almost equal to the English model. Accuracy is surprisingly even higher for the round-trip translated validation set. We hypothesize this is caused by the translation serving as a normalization of the training sentences, fixing typos and similar errors, and in our opinion actually improving the quality of some of the example sentences. See Supplemental material, section \ref{translation_examples} for examples.

 This result suggests that the inferior performance of the model trained on Czech dataset was not caused by translation errors, but rather by worse pretraining of the multilingual BERT model in Czech. To gather further insights, we did a round-trip translation of the training data back into English and trained the model on this dataset. The resulting model performs slightly worse than the model trained on the original data, but still better than training in Czech directly. We also evaluated the best single model, ALBERT-xxlarge, with the round trip translated validation set. The result is 2.7 F1 worse than on the authentic validation set. 

To quantify the effect of translation quality, we also translated the data with a popular online translator, which in our experience provides good results, yet worse than our model. We observe a performance drop of 2.1 accuracy points for the best ALBERT classifier, compared to translations made by our NMT system. Since we know our classifier detects nonsensical statements with high accuracy, we can turn this method around and it can be also of limited use to compare machine translation models -- if the detection on a round-trip translated validation set (or train set, to improve the accuracy of the classifier even further) is accurate, we can assume that the translation model  preserves the meaning of the translations.

We note that the round-trip translation setting is not equivalent to using the pipeline on authentic Czech sentences, since the Czech version of the inputs is created by machine translation a thus is less diverse and easier to translate back into English for an NMT model.

\subsection{Subtask B}
Based on experience from subtask A, we experimented only with roBERTa-large and ALBERT-xxlarge models. ALBERT-xxlarge outperforms roBERTa-large model by 1.2 accuracy points on the validation set. The results are presented in Table \ref{4b_results}.

The final submitted ensemble was created in similar fashion as in subtask A, we found a combination of 15 best single models that has the best accuracy on the dev set. Our submitted ensemble consists of 5 models and it ranked 8th in the official results, 1.9 accuracy behind the winning submission.

Our submission misclassifies 58 out of 997 validation examples. We provide  error analysis in the Supplemental material, section \ref{subtaskB_examples}.
We also assessed ability of our subtask C submission to detect and explain nonsensical statements using data from subtask B. 
Inversely, we used our subtask B submission to rerank hypotheses of our subtask C system. For more details, see the following subsection.

\begin{table}[ht!]
\begin{center}
\begin{tabular}{|l|ccc|}
\hline \bf Model & \bf dev accuracy  & \bf test accuracy & $\boldsymbol{\#\theta}$  \\ \hline
Majority & 34.43 & - & -\\ \hline

roBERTa-large & 91.78& & 355M \\ \hline

ALBERT-xxlarge &  93.0 &  92.4 & 223M \\\hline
ALBERT-xxlarge-ens* &  \textbf{94.08 } &\textbf{ 93.1} & 5*223M\\ \hline

\end{tabular}
\end{center}
\caption{Results of subtask B, accuracy on validation and test sets. Our submission is marked by an asterisk.}
\label{4b_results}
\end{table}
\subsection{Subtask C}

In subtask C, the input was a nonsensical statement and the goal was to generate a proper reason why the statement does not make sense. The systems were compared in terms of BLEU score and human ranking. Our baseline was to simply copy the source sentence and compute the BLEU score.

\paragraph{Models}
We experimented with two approaches: sequence-to-sequence model and language model predicting the next token. 
For the sequence-to-sequence approach, we compared 3 systems. First, the vanilla Transformer-base (same hyperparameters as in \newcite{vaswani2017attention}) model trained from scratch. Secondly, the same Transformer model, but pretrained on machine translation between Czech, English, Polish, Slovak and Russian and finetuned for this task. And finally, the BART model. For the language model approach, we chose to experiment with two sizes of GPT-2 model. 

\paragraph{Results} The results are presented in Table \ref{tab:subtaskC}. The BART model performed the best both in human evaluation and BLEU score. The second best system in terms of BLEU score was the finetuned NMT model, but as we discuss further, BLEU scores do not correlate well with the human judgements. We performed our own manual evaluation of the generated explanations based on this observation. We selected four systems -- our final BART submission, GPT-2-medium, NMT pretrained transformer-base and vanilla transformer-base trained on the task data only. We evaluated the first test set 100 sentences generated by each model, using a  0-3 scale described by the task organizers. In our human evaluation, GPT-2 ranked second. The finetuned NMT model learned mostly to copy the source sentence or create simple negations.
Still, in some cases, the vanilla Transformer, even without NMT pretraining, was able to generate valid explanations, where other models failed. See Supplemental material, section \ref{subtaskC_examples} for example outputs of the models.

\begin{table}[ht!]
\begin{center}
\begin{tabular}{|l|cccc|}
\hline \bf Model & \bf dev BLEU & \bf test BLEU  & \bf human eval & $\boldsymbol{\#\theta}$  \\ \hline
Copy source & 16.53 & 17.23 & & -\\ \hline

Transformer-base & 10.62 & 10.11 & 0.31 & 65M \\ \hline
Transformer-base NMT pretrained  &  16.76& 16.82 & 0.98& 65M \\  \hline

GPT-2 small & 13.03  & 13.15 & & 117M \\ \hline
GPT-2 medium & 14.34  & 16.05 & 1.36& 345M \\ \hline
BART &    19.53 & 21.07 & & 406M\\ \hline
BART + src augmented & 19.47 &  19.91 & & 406M\\ \hline
BART + src, tgt augmented & 19.59  & 20.61 & & 406M\\ \hline
BART ensemble & 21.71& 21.30 & & 406M\\ \hline
BART ensemble, tuned beam size*& \textbf{22.11} & \textbf{22.39 }& 1.62 (1.84) & 3*406M\\ \hline
reranked by task B & - & 19.22 & \textbf{1.70} & - \\ 
\hline
\end{tabular}
\caption{Validation and test set BLEU scores on task C. Human eval displays mean score we assigned to the first 100 test sentences generated by the model, based on the scale used by the organizers. The human evaluation score assigned by the organizers for our submission is shown in parentheses. Our final submission is marked by an asterisk.  The last row shows BLEU scores and human evaluation results of explanations generated by GPT-2 medium, BART ensemble and NMT pretrained Transformer, reranked by our task B submission.}
\label{tab:subtaskC}
\end{center}
\end{table}

\paragraph{Data augmentation}
We experimented with data augmentation via round-trip translation, similarly to subtask A. We machine translated the training data to Czech and back into English, and concatenated the translated examples with the original training data. We evaluated variants where only source sentences were translated, and where both source and target sentences were translated. 

\paragraph{Ensembling}
For the final submission, we evaluated all combinations of the 10 best checkpoints from all BART runs and chose the combination with the best validation set BLEU. The final ensemble consists of 3 models, one trained only on authentic training data, one trained on source augmented data and one trained on both source and target augmented data. We tuned beam size and length penalty on the validation data.

\paragraph{Metric analysis}
We suspected that the correlation between human rating and BLEU scores could be low, since BLEU score only measures token overlap. Overall correlation between BLEU score and human score in the official ranking is quite high (Pearson's $r=0.831$), because a lot of submissions scored poorly in both metrics. If we consider only the top 5 BLEU scoring systems, Pearson's correlation coefficient drops to $0.095$. Indeed, our submission won in the BLEU score rating, however, in the human rating, it was ranked 4th. On the other hand, the system  that was ranked first by the human evaluators had even lower BLEU score than our baseline, which was to copy the source sentence to the output.

As an another way to assess the ability of the models to explain why a sentence contradicts common sense, aside from human evaluation, we used data from subtask B. For each example, the three choices were scored by the model in terms of perplexity given the input statement. The choice with the lowest perplexity was selected. Since perplexity is a function of cross-entropy loss used during the training, given the input statements $s$, and the three possible explanations $o$ we can choose the best answer $y$ as follows: 
\vspace{-2pt}
$$y=\argmin_{i \in \{1,2,3\}}(loss_{o_{i}}(s))$$
\vspace{-2pt}

Resulting accuracy (see Table \ref{4b_ppl}) is more than 20 points lower than our best subtask B submission. These results suggest that although our subtask C submission ranked 1st in the BLEU evaluation, the model ability to explain why a statement is against commonsense is limited, which was proven by a human evaluation, where our system ended up 4th.

Adding a similar metric into the evaluation may increase the correlation between automated and human evaluation. In our setting, the perplexity ranking on subtask B agrees with the human ranking of the models, whereas BLEU ranking swaps second and third place (GPT-2 and NMT Transformer).

\paragraph{Reranking}
After evaluation, we saw BART model performing the best in all presented metrics. However, in many cases, GPT-2 or even the NMT pretrained Transformer model, generate better explanations, as shown in section \ref{subtaskC_examples} of the Supplemental material.

We had already established that our model from task B can choose a correct explanation why a statement is nonsensical with high accuracy. We used task C outputs of BART, GPT-2 and NMT Transformer to create a dataset in a format suitable for task B, i.e. the statement, and three possible explanations, each generated by one of the models. We used our task B submission to choose the best option from these three. An explanation generated by BART was selected in 49.8\% of examples, an explanation from GPT-2 in 31.6\% of examples and finally, Transformer-base explanation was chosen in 18.6\% of the cases.  We recomputed our human evaluation score done on the first 100 examples (since all the examples were already scored when we evaluated the systems which are being reranked, the same scores were used). The score after reranking reached 1.70, 0.08 better than the best system by itself. The upper bound of evaluation score after reranking, if the best option is chosen for each example, was 1.88.

In qualitative analysis, we saw the reranking often selected the simple negations generated by the NMT Transformer as a fall-back option in case none of the more complex models was able to generate a proper reason, e.g. they only generated a paraphrase of the input. This makes the model more robust, but it does not reflect much in the human evaluation, since, if we understood the criteria correctly, both simple negation and paraphrase of the input are scored the same.

\begin{table}[t!]
\begin{center}
\begin{tabular}{|l|cccc|}
\hline \bf Model & \bf task B dev acc  &  \bf task B acc rank &  \bf human rank &  \bf BLEU rank \\ \hline

 BART-ens  & 72.92  & 1 &  1& 1\\
BART-single no synth  & 73.92  &- &- & -\\
BART-single synth   & 73.52  &  -& -&- \\
GPT-2 &  55.57 &  2 & 2 &3\\

Transformer-base NMT  &  50.35  & 3 & 3&2\\
Transformer-base &  46.34 &  4 & 4 & 4 \\

\hline
\end{tabular}
\caption{Accuracy of models from task C when used on task B (task B options are scored by perplexity of the explanation given the statement). Perplexity ranking agrees with our human evaluation, whereas BLEU ranking swaps GPT-2 and Transformer pretrained on NMT.}
\label{4b_ppl}
\end{center}
\end{table}

\section{Related work}
Classical approach to machine commonsense reasoning is to use knowledge graphs and symbol manipulation to infer an answer to a commonsense question, or to find out if a statement agrees with common sense. One of the largest general purpose knowledge bases is ConceptNet by \newcite{speer2017conceptnet}.

With the rise of pretrained language representation models (LRM), it is being questioned whether knowledge bases and exact inference are necessary for commonsense understanding, or the LRMs already have commonsense  capabilities embedded into them. A work by \newcite{liu2020k} combines both approaches, integrating relevant parts of a knowledge graph into an input of BERT-like model. 

Authors of currently the largest pretrained language model, GPT-3 \cite{brown2020language}, provide a comprehensive analysis of the reasoning capabilities of the model on various tasks. However, whether the pretrained models really perform reasoning, or rather they only match patterns present in the training data, remains an open question.

Winograd schema challenge \cite{levesque2012winograd} and Winogrande \cite{sakaguchi2019winogrande} are benchmarks for natural language understanding and commonsense reasoning, considered an alternative to the Turing test. The task is to disambiguate a pronoun, which has an antecedent in previous statement, based on common sense. A relaxed version of the challenge is a part of  GLUE \cite{wang2018glue} natural language understanding dataset.

CommonsenseQA \cite{talmor2018commonsenseqa} is a multiple-choice dataset for commonsense question answering, consisting of questions that are difficult for current natural language processing models and require prior knowledge.

\newcite{rajani2019explain} created a crowd-sourced common sense explanation dataset, called CoS-E, and trained BERT-based model on this dataset, improving results on CommonsenseQA task.

SWAG \cite{zellers2018swag} is a dataset containing multiple choice questions about grounded situations. The question is a the start of a situation, and the task is to choose most probable continuation of the situation.
\section{Conclusions}

 We have demonstrated that pretrained language representation models are able to obtain high accuracy on datasets for subtasks A and B. 
We experimented with solving subtask A in the Czech language by means of machine translating data and training multilingual models. We have shown that using an English model and simply translating the input sentences from Czech to English yields superior performance compared to multilingual models, even though this approach is prone to error propagation.  

In subtask C, we used pretrained sequence-to-sequence and language models.
We demonstrated that BLEU score does not correlate well with human evaluation in this subtask, and we propose a perplexity metric based on the subtask B dataset, which agrees more with our manual ranking. Finally, we show that reranking hypotheses of subtask C systems by our subtask B model improves the manual ranking score.

\section*{Acknowledgements}

This work was supported by the Czech Ministry of Education, Youth and Sports, subprogram INTERCOST, project code: LTC18054.

\bibliographystyle{coling}
\bibliography{semeval2020}

\appendix

\label{sec:supplemental}
\section{Error breakdown}
In this section, we analyse errors made by our best system on validation set of each subtask.
\subsection{Subtask A}
\label{subtaskA_examples}
Ensemble of ALBERT-xxlarge models chose wrong label in 28 (2.81\%) out of 997 validation examples. While most of the misclassified examples were really errors made by the model, we consider some of the mistakes wrongly labeled or at least debatable. First, a few examples of mistakes that are definitely a model error a not a data problem (label denotes which statement is against common sense):

\vspace{5pt}
\begin{tabular}{p{0.4\linewidth}p{0.35\linewidth}cc} 
Statement 0 & Statement 1 & Pred & Label \\ \hline
    The largest animal on the land is the elephant & The largest animal is the elephant & 0 &1 \\  
    an equilateral triangle may have three unequal sides & an equilateral triangle has angles of 60, 60, and 60 degrees respectively & 1 &0 \\
 \hline

    \end{tabular}
    
\vspace{10pt}
    
Oddly, it seems that the model is having a hard time grasping facts about reproduction and birth. There are 5 examples with words birth or born in the validation set. 4 of them have different predicted and reference labels:

\vspace{5pt}

\begin{tabular}{p{0.4\linewidth}p{0.35\linewidth}cc} 
Statement 0 & Statement 1 & Pred & Label \\ \hline
A chicken lays eggs when it gives birth. & A chicken gives birth to live chicks that are not encased in eggs.& 0 & 1 \\
The horse gave birth to a little horse & The horse gave birth to a puppy & 0 & 1 \\
the mother gave birth to a baby boy & the father gave birth to a baby boy & 1 & 1 \\
babies are born naked & babies are born with clothes on & 0 & 1  \\

Some mammals are born with two paws & Every mammal is born with two paws & 1 & 0 \\
 \hline

\end{tabular}
\vspace{10pt}

However, the last example is classified correctly by the model and the label is incorrect, since we can easily think of a four pawed mammal. This is the only label in the 28 wrongly classified examples we completely disagree with.
however, in some of the misclassified examples, neither of the sentences are strictly against common sense:

\vspace{5pt}

\begin{tabular}{p{0.4\linewidth}p{0.35\linewidth}cc} 
Statement 0 & Statement 1 & Pred & Label \\ \hline

the girl on the phone is annoying & the phone is annoying & 0 & 1 \\
It was late, so she hurried up & It was late, so she stopped and had a rest & 0 &1 \\
 everyone in the world needs a helmet & everyone needs a heart &  0 & 1 \\
 Not every item in the grocery store is taxable.& Each item in the grocery store is taxable. &0 & 1 \\
  \hline

\end{tabular}

\vspace{5pt}

Both people talking over a phone and phones themselves can be annoying at times.  A late time of a day can be a good reason to stop and rest. As for the third example, a man lived for 555 day without a heart waiting for a transplant\footnote{\url{https://healthblog.uofmhealth.org/heart-health/living-for-years-without-a-heart-now-possible}}.
The correct label of the last example depends on whether you live in Texas, where, according to a government document\footnote{\url{https://comptroller.texas.gov/taxes/publications/96-280.pdf}}, food items in grocery stores are not taxable.

Finally, in this example, both statements are against common sense:

\vspace{5pt}

\begin{tabular}{p{0.4\linewidth}p{0.35\linewidth}cc} 
Statement 0 & Statement 1 & Pred & Label \\ \hline

A bald man brushed his hair every day & A bald man washed his hair every day & 1 & 0
 \\ \hline

\end{tabular}

\vspace{10pt}

\subsection{Subtask B}
\label{subtaskB_examples}
Our submission classified 58 (5.81\%) out of 997 validation set examples into different class than the gold label. We analysed the differently classified examples and in 41 cases, we agree with the data label and these cases are clearly a model mistake. In 17 cases, we consider the example problematic. First, we present a few examples of cases were the example is good quality and the model prediction is wrong:

\vspace{5pt}

\begin{tabular}{p{0.25\linewidth}p{.52\linewidth}cc} 
Statement & Options &  Pred & Label \\ \hline

She put the freezer in the microwave.& \makecell[ll]{\textbf{A:} Freezers are full of frozen items. \\ \textbf{B:}  Freezers are usually bigger than the microwaves \\\textbf{C:} Microwave is a type of oven that cooks or heats \\  food very quickly. } & C & B
 \\ \\
 
 I screamed when I hit my foot against the pillow.& \makecell[ll]{\textbf{A:} A pillow is used for sleep.\\ \textbf{B:} A pillow is soft.\\\textbf{C:} You can hit your foot on things.}&A&B \\ \\
 
 Homestar Runner is Lazy and can't run & \makecell[ll]{\textbf{A:} Homestar runner has no hands so therefore he  \\ cannot participate in sports \\ \textbf{B:}  Homestar runners body is mostly legs and he \\ runs everywhere\\ \textbf{C:}  Homestar Runner doesn't exist so he cannot be an \\  athlete} &C &   B \\

 \hline
\end{tabular}
 \vspace{5pt}

 In two examples, we found typos in the statement or in one of the options: 
 \vspace{5pt}

\begin{tabular}{p{0.25\linewidth}p{.52\linewidth}cc} 
Statement & Options &  Pred & Label \\ \hline

Sunlight and water help cars grow.& \makecell[ll]{\textbf{A:} Cars have already reached their full size. \\ \textbf{B:}  Sun heats a car too much. \\\textbf{C:} Cars are ma\textcolor{red}{d} made. } & A & C \\ \\
Tofu is a ufo substitute & \makecell[ll]{\textbf{A:} Ufot is unhealthy \\ \textbf{B:} Tofu is real \\\textbf{C:} Ufot is something I made up}&A&C
 \\
  \hline

\end{tabular}
 \vspace{10pt}

In some cases, the statement is not against common sense:

\vspace{5pt}

\begin{tabular}{p{0.25\linewidth}p{.52\linewidth}cc} 
Statement & Options &  Pred & Label \\ \hline
cat watches tv\footnotemark& \makecell[ll]{\textbf{A:} cat can walk around and t.v does not walk \\ \textbf{B:} cats can not watch t.v \\ \textbf{C:} cat relaxes in front of tv}& B&C \\ \\

he walked in front of a car & \makecell[ll]{\textbf{A:}  the cars were waiting at the pedestrian light \\ for him to cross the road \\ \textbf{B:} pedestrain has to wait until it is safe to cross \\ the road \\ \textbf{C:} its hard to see a pedestrian at night if you \\ have bad vision}& B&A \\  
\hline
\end{tabular}
 
 \vspace{10pt}
 
\footnotetext{{Cats do watch television \url{https://www.youtube.com/watch?v=U6PXDdwhhqM}}}
For some of the examples, we believe the model prediction is at least as correct as the label, for instance:
 \vspace{10pt}

\begin{tabular}{p{0.25\linewidth}p{.52\linewidth}cc} 
Statement & Options &  Pred & Label \\ \hline
An apple can voice an opinion & \makecell[ll]{\textbf{A:}  An apple has no mouth \\ \textbf{B:}  An apple has no legs while a person has two legs \\ \textbf{C:}  An apple cannot speak}&C&A \\ \\

today I went to the Mars& \makecell[ll]{\textbf{A:}  No one has ever walked on the Mars\\ \textbf{B:} Mars is a star in the universe \\ \textbf{C: }The Mars is hot}&A&B   
\\ 
\\

You go to a concert for the lectures.& \makecell[ll]{\textbf{A:} People go to concerts to have fun. \\ \textbf{B:}  Some people dance at concerts. \\ \textbf{C:} Music bands are on the stage at a concert}& A&C \\ \\

The color of the grass is purple.& \makecell[ll]{\textbf{A:} Eggplants are purple and the grass is not the \\ same color as an eggplant.\\ \textbf{B:} Grass is not the same shape as eggplants \\ \textbf{C:} Chlorophyll absorbs blue and red light but does \\ not produce purple pigment in grass.} & C & A \\

  \hline
\end{tabular}
 
 \vspace{10pt}
 
Finally, we believe that in some examples which were classified incorrectly by our model, none of the labels are correct:
 \vspace{5pt}

\begin{tabular}{p{0.25\linewidth}p{.52\linewidth}cc} 
Statement & Options &  Pred & Label \\ \hline
the basketball loves to play Steven& \makecell[ll]{\textbf{A:} Steven loves to eat unhealthy foods rather than playing basketball\\ \textbf{B:} Steven wants to play with his brother basketball game \\ \textbf{C:} Steven don't play basketball} & C  & B \\
\\

 He walked through the floor.& \makecell[ll]{\textbf{A:} Gravity makes people down. \\ \textbf{B:}  The floor was too thick. \\ \textbf{C:}  The floor was too tall.}&B&A   \\
\\

Trees can not live on the ground & \makecell[ll]{\textbf{A:} Trees need water to live \\ \textbf{B:}  some kinds of trees can also live in the water  \\ \textbf{C:}  Human also need water and salt to live} & A & B \\

  \hline
\end{tabular}

 \vspace{10pt}
\subsection{Subtask C}
\label{subtaskC_examples}

We chose 15 examples from the first 100 examples from the test set and we present explanations generated by the systems we have experimented with, sorted from the best to the worst system (1=BART ensemble, 2=GPT-2 medium, 3=NMT-pretrained Transformer, 4=vanilla Transformer). The explanation scored best by task B system reranker is marked by an asterisk, Transformer without pretraining was not among the ranked systems.

\begin{tabular}{p{0.4\linewidth}p{.55\linewidth}} 
Statement & Explanations  \\ \hline
the family adopted a dinosaur to be their new pet & \makecell[ll]{\textbf{1*:} Dinosaurs are extinct.  \\ \textbf{2*:} Dinosaurs are extinct.    \\ \textbf{3:}  Dindinosaur is not a pet  \\ \textbf{4*:} Dinosaurs are extinct.} \\  \hline

A normal person has 4 arms & \makecell[ll]{\textbf{1:} A normal person has two arms  \\ \textbf{2*:}  A normal person has only 2 arms.  \\ \textbf{3:} A normal person has 4 arms  \\ \textbf{4:}A person has only two arms. } \\  \hline

"in summers, always the weather is cold." & \makecell[ll]{\textbf{1:} Summer is the hottest season of the year  \\ \textbf{2:} Summer is not cold.  \\ \textbf{3*:}   Summer is the hottest season of the year. \\ \textbf{4:} Summer weather is cold.} \\  \hline

The stone can float on the water & \makecell[ll]{\textbf{1*:} Stones do not float.  \\ \textbf{2:}  Stones are not water. \\ \textbf{3:}  Stones cannot floon on water \\ \textbf{4:} Stones are too heavy to float in water.} \\ \hline

You should leave the lights on when you leave the room & \makecell[ll]{\textbf{1*:} You should turn the lights off when you leave the room.  \\ \textbf{2:}  Lights are not used when you leave the room.  \\ \textbf{3:} Lights should be on when you leave the room.   \\ \textbf{4:} You should leave the lights when you leave the room. } \\ \hline

Rabbits are carnivores & \makecell[ll]{\textbf{1*:} Rabbits are herbivores.  \\ \textbf{2*:} Rabbits are herbivores.  \\ \textbf{3:}  Rabrabrabrabis carnivores  \\ \textbf{4:} Rabbits are carnivores.} \\ \hline

The chef put extra lemons on the pizza. & \makecell[ll]{\textbf{1: } Lemons are not edible.  \\ \textbf{2*:} A chef would not put extra lemons on a pizza.   \\ \textbf{3:}  Pizza is not a place to put extra lemons.  \\ \textbf{4:} Lemons do not sell food.} \\ \hline

She went to the grocery store to get an aneurysm. & \makecell[ll]{\textbf{1*:} Grocery stores don't have aneurysms. \\ \textbf{2:} Grocery stores don't have an aneurysm.   \\ \textbf{3:} She went to the grocery store to get an anew  \\ \textbf{4:} Grocery stores do not sell grocery stores.} \\ \hline

If you're poor you can buy a yacht. & \makecell[ll]{\textbf{1:} Yachts are expensive.  \\ \textbf{2:} A yacht is not a means of transportation.  \\ \textbf{3*:} You can't buy a yacht if you are poor.  \\ \textbf{4:} Poor people don't need to feel happy } \\ \hline

The inverter was able to power the continent. & \makecell[ll]{\textbf{1:} An inverter cannot power a continent.  \\ \textbf{2*:} An inverter is not able to power the continent.  \\ \textbf{3:} The inverter is unable to power the continent.  \\ \textbf{4:} A snake cannot fly. } \\ \hline
\end{tabular}

\begin{tabular}{p{0.4\linewidth}p{.55\linewidth}} 
Statement & Explanations  \\ \hline
The branch ate the parrot. & \makecell[ll]{\textbf{1*:} A branch is an inanimate object.  \\ \textbf{2:} A branch is not a food.  \\ \textbf{3:} A branch cannot eat a parrot.   \\ \textbf{4:} Parrots are not edible. } \\ \hline

I put on my sunglasses because it was an overcast day.  & \makecell[ll]{\textbf{1:} Sunglasses don't protect you from the sun  \\ \textbf{2*:} Sunglasses are not used to protect the eyes from the sun.   \\ \textbf{3:}  Sunglasses are too small to be put on.  \\ \textbf{4:} Sunglasses do not protect sunglasses from sunglasses.} \\ \hline

Kim is dating the garbage. & \makecell[ll]{\textbf{1:} Garbage is not a person  \\ \textbf{2*:}  Garbage is not a person.  \\ \textbf{3:} Garbage cannot be dated.  \\ \textbf{4:} Garbage is not alive } \\ \hline

A television plays audio only. & \makecell[ll]{\textbf{1*:} A television does not play audio.  \\ \textbf{2:} A television can play video.  \\ \textbf{3:} Audio only is played by a television.  \\ \textbf{4:} Television is not a television } \\ \hline

God lives physically on earth & \makecell[ll]{\textbf{1:} Humans cannot live physically on earth.  \\ \textbf{2*:}  No one can live physically on earth. \\ \textbf{3:} There is no god on earth   \\ \textbf{4:} God has been seen in space. } \\ 

  \hline
\end{tabular}

\section{Translations}
\label{translation_examples}
While experimenting with the task in the Czech language, we experienced a number of issues related to translation of the data. Overall, the translation quality was good, from the first 100 examples, only two sentences were translated incorrectly. First, we show some of the good quality translations:

\vspace{10pt}

\begin{tabular}{p{0.1\linewidth}p{0.75\linewidth}} 
    Source: & His father is a wealthy businessman so he never worries about the money	\\ 
     Translation: & Jeho otec je bohatý obchodník, takže si s penězi nikdy nedělá starosti.  \\ \\
   Source: & A walk-in closet is larger than a normal closet.	\\
    Translation: &  Vestavná skříň je větší než normální skříň. \\ \\
	 Source: & He lost his phone in the mouth of a griffin.	\\ 
	 Translation: & Ztratil telefon v ústech gryfa. \\ \\ 
	 
	 Source: & I felt the building was shaking and then I realized it was an earthquake	\\
	 Translation: &  Cítil jsem, jak se budova třese, a pak jsem si uvědomil, že je to zemětřesení. \\
                 
\end{tabular}

\paragraph{Cultural context}
Some commonsense knowledge examples in the data are culturally dependent, for example this statement: 
\begin{displayquote}
\textit{Passing your driving license exams requires studying for your classes.
}\end{displayquote}

Is labeled as against common sense in the data. Such label may be correct in English, under assumption that the statement concerns situation in USA. However, if it is translated into Czech, and Czech cultural context is assumed, the statement is perfectly fine, since in Czech Republic theoretical classes are mandatory for passing the driving license exam.
\paragraph{Rare words}
Translation of rare words and expressions is a well known issue in NMT, and it has surfaced during our experiments. For instance: 

\vspace{10pt}
\begin{tabular}{p{0.1\linewidth}p{0.75\linewidth}}
    Source: & Perming will make your hair longer. \\ 
     Translation: & \textcolor{red}{Perming} vám prodlouží vlasy.  \\ \\
               Source: & She put mustard on a corndog.	\\
               Translation: & Dala hořčici na \textcolor{red}{kukuřičného psa.} \\ \\

                Source: &  They put on ghillie suits to go outside \\
                Translation: & \ Oblékají si \textcolor{red}{šilinkové obleky}, aby šli ven.
                
\end{tabular}

In the first case, the verb perming is not translated at all. This word does not exist in Czech language, and to our knowledge, it needs to be translated periphrastically, as Czech does not have a synonymous verb.

In the second case, corndog is translated literally -- dog as in animal and corn as an adjective meaning made from corn. These two sentences are the only completely wrong translations observed in the first 100 training examples.

Finally, in the third case, \textit{ghillie suits} is translated as \textit{shilling suits}, which doesn't make any sense.
\paragraph{High perplexity}

Rarely, a simple sentence which is against the common sense is translated incorrectly. We hypothesize that translation of such nonsensical statements might be less adequate in regard to the source sentence, since higher perplexity of these statements in the target language may cause the model to generate less adequate, but more probable target sentences. This may be caused by the language modelling (LM) part of the translation --  even though the translation is not difficult, the sentence generated by the model would be so improbable in the target language, that a wrongly translated, but more probable sentence is generated instead, i.e. that the nonsensical sentences have high LM perplexity and the  LM component of the model forces generation of sensible statements without regard for the source sentence.  For instance, in this case, the model translates the sentence as if the subject was feminine:

\vspace{10pt}

\begin{tabular}{p{0.1\linewidth}p{0.75\linewidth}}
    Source: & He gave birth to a baby. \\ 
     Translation: & Porodil\textcolor{red}{a} dítě. \\
\end{tabular}

\paragraph{Round-trip translations}
In some of the experiments, we translated the validation and test sets to Czech and back into English. Even thought translating twice is prone to error propagation, we did not see many wrong translations. In some cases, the translated sentences were even better than the original ones, since the NMT model was robust enough to correctly translate inputs with typos, wrong tenses or casing and similar minor issues, which were present in the data. Few examples of these round trip translations are shown below:

\vspace{10pt}

\begin{tabular}{p{0.1\linewidth}p{0.75\linewidth}}
    Source: & Boys that play pee wee football follow strict hitting guidelines and rules. \\
    Translation: & Guys who play pee football follow strict rules and guidelines. \\ \\

 Source: & since he is good man police jailed him \\
 Translation: & Because he's a good man, the police imprisoned him. \\ \\

 Source: &  He spends \$100 for dinner.\\
  Translation: &  He spends \$100 on dinner.\\

\end{tabular}
\newpage
\section{Hyperparameters}
In this part, we present training parameters of our submitted models. If a range of number is given, we searched for the optimal value in this range. 
\subsection{Subtask A and B}
\begin{table}[ht!]
\centering
\begin{tabular}{|c|c|}
\hline
\textbf{Hyperparameter}         & \textbf{Value} \\ \hline
Max gradient norm       &    1.0     \\ \hline
    Epochs         &         6 \\ \hline
    Dropout & 0.0  \\ \hline 
         Batch size & \numrange[range-phrase = -- ]{32 }{ 72 } \\ \hline
        Learning rate & \numrange[range-phrase = -- ]{5e-6 }{ 4e-5 }  \\ \hline

    Optimizer & Adam ($\epsilon$ = 1e-8) \\ \hline
\end{tabular}
\caption{Hyperparameters for Subtasks A and B, ALBERT model. }
\end{table}

\subsection{Subtask C}
\begin{table}[ht!]
\centering
\begin{tabular}{|c|c|}
\hline
\textbf{Hyperparameter}         & \textbf{Value} \\ \hline
Max gradient norm       &    0.1     \\ \hline
    Epochs         &         4 \\ \hline
    Dropout & 0.0  \\ \hline 
         Batch size & \numrange[range-phrase = -- ]{1 }{ 16 } \\ \hline
        Learning rate & \numrange[range-phrase = -- ]{5e-6 }{ 2e-5 }  \\ \hline

    Optimizer & Adam ($\epsilon$ = 1e-8) \\ \hline
\end{tabular}
\caption{Hyperparameters for Subtasks A and B, ALBERT model. }
\end{table}

\end{document}